\pdfoutput=1
\documentclass[11pt]{article}

\usepackage[preprint]{acl}
\usepackage{times}
\usepackage{latexsym}
\usepackage[T1]{fontenc}
\usepackage[utf8]{inputenc}
\usepackage{microtype}
\usepackage{inconsolata}
\usepackage{graphicx}

\usepackage{booktabs} % for table
\usepackage{tabularx} % for table
\usepackage{tcolorbox} % for prompt template
\usepackage{soul} % for highlight effect
\usepackage{enumitem} % for bullets
\usepackage{cleveref} % for section symbol (cref)
\crefformat{section}{\S#2#1#3} % see manual of cleveref, section 8.2.1
\crefformat{subsection}{\S#2#1#3}
\crefformat{subsubsection}{\S#2#1#3}

\newcommand{\method}{REFIND}
\title{REFIND at SemEval-2025 Task 3: Retrieval-Augmented \\ Factuality Hallucination Detection in Large Language Models}

\author{
    DongGeon Lee, Hwanjo Yu\thanks{Corresponding author} \\
        Pohang University of Science and Technology (POSTECH) \\
        Pohang, Republic of Korea \\
    \texttt{\{donggeonlee, hwanjoyu\}@postech.ac.kr}
}

\begin{document}
\maketitle

\begin{abstract}

Hallucinations in large language model (LLM) outputs severely limit their reliability in knowledge-intensive tasks such as question answering.  
To address this challenge, we introduce REFIND (Retrieval-augmented Factuality hallucINation Detection), a novel framework that detects hallucinated spans within LLM outputs by directly leveraging retrieved documents.
As part of the REFIND, we propose the \textit{Context Sensitivity Ratio (CSR)}, a novel metric that quantifies the sensitivity of LLM outputs to retrieved evidence. This innovative approach enables REFIND to efficiently and accurately detect hallucinations, setting it apart from existing methods.
In the evaluation, REFIND demonstrated robustness across nine languages, including low-resource settings, and significantly outperformed baseline models, achieving superior IoU scores in identifying hallucinated spans.
This work highlights the effectiveness of quantifying context sensitivity for hallucination detection, thereby paving the way for more reliable and trustworthy LLM applications across diverse languages. Our code is available at \url{https://github.com/oneonlee/REFIND}.

\end{abstract}

%%%%%%%%%%%%%%% Summary
% REFIND is a retrieval-augmented framework for detecting hallucinated spans in LLM outputs by leveraging retrieved documents. It introduces Context Sensitivity Ratio, a metric quantifying LLM sensitivity to evidence. REFIND outperforms baselines across nine languages, including low-resource settings, achieving superior hallucination detection accuracy. These results demonstrate the effectiveness of context sensitivity quantification in improving hallucination detection.

\section{Introduction}

Detecting hallucinated information in responses generated by large language models (LLMs) has emerged as a critical challenge in the field of natural language generation \cite{Ji2023HallucinationSurvey, Zhang2023Sirens_Song_Hallucination}. 
Hallucination, in this context, refers to the generation of content that is factually incorrect or lacks grounding in verifiable sources \cite{li-etal-2024-dawn}. 
This issue is particularly pronounced in knowledge-intensive tasks that demand high factual accuracy, such as question answering \cite{Lee2022Factuality, sun-etal-2024-towards-verifiable}.
The consequences of unmitigated hallucination are significant, ranging from the propagation of misinformation to a decline in trust in AI systems, underscoring the need for effective hallucination detection for the development of safe and trustworthy AI.

\begin{figure}[t!]
    \centering
    \includegraphics[width=\columnwidth]{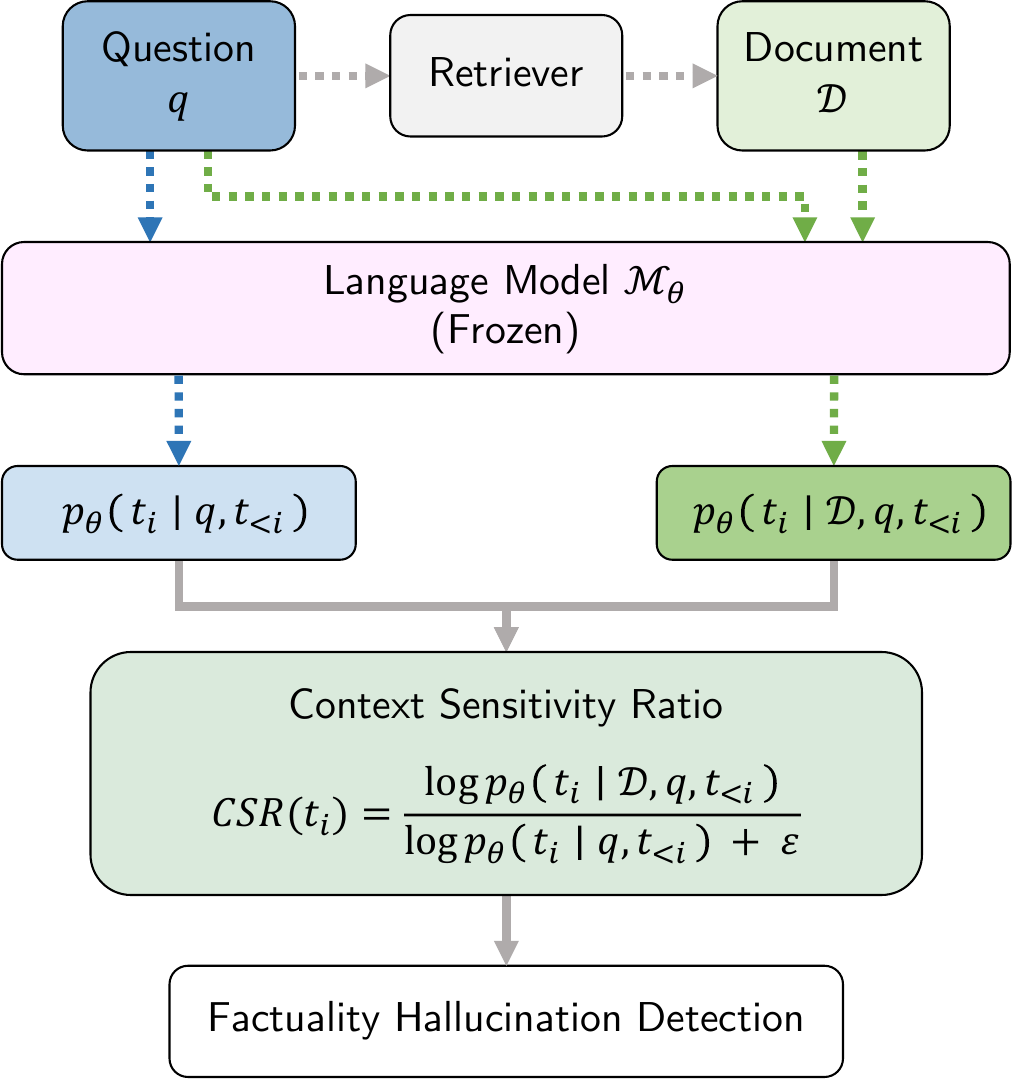}
    \caption {
        An overview of the proposed \textbf{\method} method.
        (1) Given a question $q$, a set of relevant documents $\mathcal{D}$ is retrieved using a retriever $\mathcal{R}$.
        (2) A frozen language model $\mathcal{M}_\theta$ computes token probabilities $p_{\theta}(t_i \mid \cdot )$ for each token $t_i$, with and without the retrieved context $\mathcal{D}$.
        (3) The Context Sensitivity Ratio (CSR) is calculated for each token $t_i$. Tokens with the CSR exceeding a predefined threshold $\delta$ are classified as hallucinations.
    } 
    \label{fig:overview}
\end{figure}

Prior research has explored various approaches for hallucination detection. 
Token-level classifiers, for example, leveraging pre-trained language models like RoBERTa \cite{Liu2019RoBERTa}, have been employed for binary classification, labeling individual tokens as either factual or hallucinated \cite{liu-etal-2022-token}. 
However, these models often exhibit limitations when applied to low-resource languages and tend to rely heavily on internal knowledge without effectively utilizing external evidence, which can hinder their performance. 
Extrinsic methods, such as retrieval-augmented models, aim to mitigate hallucinations by integrating external knowledge.  
Nevertheless, existing retrieval-augmented approaches, such as FAVA \cite{mishra2024finegrained-FAVA}, can potentially lead to inaccuracies in aligning the modified responses with the original LLM output, due to their multi-step processes involving retrieval, comparison, and editing.

To address these limitations, we introduce \textbf{REFIND} (\textbf{RE}trieval-augmented \textbf{F}actuality halluc\textbf{IN}ation \textbf{D}etection), a novel framework specifically designed to identify hallucinated spans within LLM-generated text.  REFIND achieves this by quantifying the context sensitivity of each token at the token level.  
By leveraging retrieved documents, REFIND calculates a Context Sensitivity Ratio (CSR) for each token in the LLM's response, measuring the token's dependence on external contextual information. 
Tokens exhibiting high CSR values are identified as likely hallucinations, offering a more direct and efficient approach to factuality verification.

Our contributions can be summarized as follows:

\vspace{-0.075in}

\begin{itemize}[itemsep=0.3mm, parsep=1pt, leftmargin=*]

    \item We present REFIND, a novel framework for detecting hallucinated spans in LLM responses by leveraging an external retriever and calculating the CSR at the token level.
    % \item We present REFIND, a novel framework for detecting hallucinations in LLM outputs by leveraging an external retriever and calculating the Context Sensitivity Ratio (CSR) at the token level.
    \item We conduct a comprehensive evaluation of REFIND using the SemEval 2025 Task 3: Mu-SHROOM dataset \cite{vazquez-etal-2025-mu-shroom}, a multilingual benchmark for detecting hallucinated spans.  REFIND is rigorously tested across nine diverse languages – Arabic, Czech, German, Spanish, Basque, Finnish, French, Italian, and English – demonstrating its robustness in both high- and low-resource settings.
    \item Experimental results demonstrate that REFIND significantly outperforms baseline models such as token-level classifiers and FAVA, achieving superior Intersection-over-Union (IoU) scores. This highlights the efficacy of the CSR in accurately identifying hallucinated content.
\end{itemize}
\section{Related Work}

\paragraph{Detection of Hallucinated Responses} Several studies have proposed methods to detect whether a response contains hallucinated information. \citet{Farquhar2024Detecting, han2024semantic, arteaga2024hallucination} leveraged semantic entropy \cite{kuhn2023semantic} to estimate uncertainty and identify hallucinations. These approaches utilize entropy-based metrics to assess the reliability of generated responses.
SelfCheckGPT \cite{manakul-etal-2023-selfcheckgpt} introduces a method that employs the language model itself to sample multiple responses and detect inconsistencies among them, thus identifying hallucinated outputs. However, this method relies solely on the internal knowledge of the language model, making it less effective when the model's knowledge is limited or incomplete.

\paragraph{Detection of Hallucinated Spans} Beyond identifying whether a response is hallucinated, other works aim to detect specific spans of hallucinated content within a response of LLMs. Token-level classification approaches \cite{liu-etal-2022-token} utilized pre-trained language models to classify individual tokens as factual or hallucinated. These methods focus on analyzing attention patterns, demonstrating that query input tokens (defined as constraint tokens) exhibit strong correlations with factual answer tokens \cite{yuksekgonul2024attention}.

FAVA \cite{mishra2024finegrained-FAVA} proposes a retrieval-augmented pipeline that integrates retrieval, comparison, and editing steps to identify and correct hallucinated spans. While effective, the multi-step process introduces complexity and alignment challenges, particularly in ensuring that the corrected responses remain consistent with the semantics of the original output.

% In summary, prior researches have laid a solid foundation for detecting hallucinated responses and spans. However, challenges such as reliance on internal model knowledge, computational complexity, and limited applicability to multilingual or low-resource settings remain. Our work builds upon these insights by proposing a novel framework that addresses these gaps through token-level context sensitivity analysis.

\section{Method}

\subsection{Task Description}
The SemEval 2025 Task 3: Mu-SHROOM \cite{vazquez-etal-2025-mu-shroom} focuses on detecting hallucinated spans in responses generated by LLMs. 
Given an input question $q$ and its corresponding LLM-generated response (along with the model's identifier), the goal is to identify spans in the response that are hallucinated. Details of the Mu-SHROOM dataset are provided in Section \ref{sec:dataset}.

\subsection{Retrieval-Augmented Factuality Hallucination Detection}

To address the challenge of factual hallucination detection in LLM outputs, we introduce \textbf{REFIND} (\textbf{RE}trieval-augmented \textbf{F}actuality halluc\textbf{IN}ation \textbf{D}etection). 
The overall workflow of the REFIND method is illustrated in Figure \ref{fig:overview}. REFIND leverages external knowledge retrieved from a relevant document set to assess the context sensitivity of each generated token.

The core principle behind REFIND is to quantify the influence of external context on the token generation process.  
We do this by measuring the change in the conditional probability of generating a token as information from retrieved documents is incorporated. 
This change is captured by the Context Sensitivity Ratio (CSR).
It quantifies the degree to which the conditional probability of generating a token is altered by the inclusion of external contextual information from retrieved documents. 

Let $\mathcal{M}_{\theta}$ denote an LLM parameterized by $\theta$, $q$ represent the input question, and $t_i$ denote the $i$-th token in the LLM's response to $q$. 
We use $p_{\theta}(t_i \mid \cdot)$ to represent the probability of generating token $t_i$ given the input.  
Furthermore, let $\mathcal{R}$ be a retriever that provides relevant documents based on $q$, and let $\mathcal{D} = \mathcal{R}(q)$ be the set of retrieved documents. 
The CSR for each token $t_i$ is defined as:
\begin{equation}
    CSR(t_i) = \frac{\log p_{\theta}(t_i \mid \mathcal{D}, q, t_{<i})}{\log p_{\theta}(t_i \mid q, t_{<i}) + \varepsilon}
\end{equation}
where $t_{<i}$ represents the sequence of preceding tokens. The numerator computes the log-probability of generating $t_i$ conditioned on the question $q$, the preceding tokens $t_{<i}$, and the retrieved document set $\mathcal{D}$. The denominator computes the log-probability of generating $t_i$ conditioned solely on the question $q$ and preceding tokens $t_{<i}$, excluding the retrieved documents.\footnote{To prevent division by zero, we use a small constant $\varepsilon$, which is set to $10^{-8}$.}

By comparing these two probabilities, the CSR effectively quantifies the sensitivity of $t_i$ to the external context provided by the $\mathcal{D}$. A higher CSR indicates a stronger influence of the retrieved context on the generation of the token.

Finally, to determine whether a token is a hallucination, we compare its CSR value to a predefined threshold, denoted as $\delta$. If the CSR value for the given token $t_i$ is greater than or equal to the threshold $\delta$, we classify that the token as a hallucination. Conversely, if the CSR value is less than $\delta$, the token is not considered a hallucination. This threshold $\delta$ serves as a hyperparameter that can be tuned to optimize the balance between precision and recall in hallucination detection.

\begin{table*}[ht]
    \centering
    \resizebox{\textwidth}{!}{
    \begin{tabular}{@{}l|ccccccccc|c@{}}
        \toprule
        Method       & AR              & CS              & DE              & EN              & ES              & EU              & FI              & FR              & IT              & \multicolumn{1}{c}{Average} \\ 
        \midrule
        XLM-R        & 0.0418          & 0.0957          & 0.0318          & 0.0310          & 0.0724          & 0.0208          & 0.0042          & 0.0022          & 0.0104          & 0.0345                      \\
        FAVA         & 0.2168          & 0.2353          & \textbf{0.3862} & 0.2812          & \textbf{0.2348} & 0.3869          & 0.2300          & 0.2120          & \textbf{0.3255} & 0.2787                      \\
        REFIND       & \textbf{0.3743} & \textbf{0.2761} & 0.3518          & \textbf{0.3525} & 0.2152          & \textbf{0.4074} & \textbf{0.5061} & \textbf{0.4734} & 0.3127          & \textbf{0.3633}             \\ 
        \bottomrule
    \end{tabular}
    }
    \caption{\label{tab:result} 
        Evaluation results on the Mu-SHROOM dataset \cite{vazquez-etal-2025-mu-shroom} using the IoU metric across eight languages: Arabic (AR), Czech (CS), German (DE), English (EN), Spanish (ES), Basque (EU), Finnish (FI), French (FR), and Italian (IT).
        The proposed method, REFIND, achieves the highest average IoU score, outperforming the baselines XLM-R and FAVA in most languages, demonstrating its effectiveness for multilingual hallucination detection.
    }

\end{table*}

\section{Experimental Setup}

\subsection{Dataset}
\label{sec:dataset}

We conduct our experiments on the Mu-SHROOM dataset \cite{vazquez-etal-2025-mu-shroom}, which consists of outputs generated by various LLMs in response to specific input questions. 
Each output is annotated by human annotators to identify spans that correspond to hallucinations.

The dataset includes multiple languages, and for our study, we focus on the following nine languages:  Arabic (AR), Czech (CS), German (DE), English (EN), Spanish (ES), Basque (EU), Finnish (FI), French (FR), and Italian (IT).
This multilingual diversity enables a comprehensive evaluation of our method across diverse linguistic contexts.

Each data point in the dataset contains the language identifier, the input question posed to the LLM, the model name, the generated output text, and its token-level probabilities. 
Additionally, binary annotations specify the start and end indices of hallucinated spans, marking each such span as a hallucination.

\subsection{Evaluation Metric}

To evaluate the performance of our hallucination detection method, we adopt the IoU metric, a standard measure for span-based evaluation.

Given the set of character indices predicted as hallucinations, $\mathcal{H}_{pred}$, and the set of character indices labeled as hallucinations in the gold reference, $\mathcal{H}_{gold}$, the IoU is calculated as:

\begin{equation}
\mathrm{IoU} = \frac{|\mathcal{H}_{pred} \cap \mathcal{H}_{gold}|}{|\mathcal{H}_{pred} \cup \mathcal{H}_{gold}|}
\end{equation}

This metric quantifies the overlap between the predicted and ground truth hallucinated spans. To handle cases where both $\mathcal{H}_{pred}$ and $\mathcal{H}_{gold}$ are empty (i.e., no hallucinations are present in either prediction or reference), we define $\mathrm{IoU} = 1.0$ to signify perfect agreement.

\subsection{Baseline Models}

\paragraph{Token-level Hallucination Classifier (XLM-R)}
We employ a token-level hallucination classifier \cite{liu-etal-2022-token} based on XLM-RoBERTa (XLM-R) \cite{conneau-etal-2020-unsupervised}, a multilingual transformer model. 
The model is fine-tuned to perform binary classification at the token level, where each token is labeled as either hallucinated or non-hallucinated. 
% By leveraging XLM-R's pre-trained multilingual representations, the classifier can effectively capture semantic discrepancies across different languages. 

\paragraph{FAVA}

We also include FAVA \cite{mishra2024finegrained-FAVA} as a baseline model. 
FAVA is a retrieval-augmented language model designed to detect and correct hallucinations in outputs generated by LLMs. 
The model is built upon Llama2-Chat 7B \cite{Touvron2023Llama2} and employs a two-step process: retrieval and editing.
To detect hallucinations in text, we compare the edited text produced by FAVA with the original text and get the span of $\mathcal{H}_{pred}$.

\subsection{Implementation Details}

The retriever $\mathcal{R}$ used to retrieve context for REFIND and FAVA employs a hybrid approach, combining sparse and dense retrieval methods.  Initially, a Wikipedia corpus is preprocessed for each language, including chunking, to serve as the retrieval corpus. The retriever first retrieves the top 10 relevant documents using BM25 \cite{Robertson2009BM25}. Subsequently, a document reranking step is performed using a pre-trained language model to select the final 5 documents to $\mathcal{D}$. To maintain consistency across the multilingual setting, we utilize \texttt{multilingual-e5-large}\footnote{\url{https://huggingface.co/intfloat/multilingual-e5-large}} \cite{Wang2024multilingual-e5} for the reranking process.

When calculating $p_{\theta}(t_i \mid q, t_{<i})$ in REFIND, we utilize the token probabilities of the LLM's output response provided in the Mu-SHROOM dataset. The computation of $p_{\theta}(t_i \mid \mathcal{D}, q, t_{<i})$ is performed using PyTorch 2 \cite{Jason2024PyTorch2}. The specific prompt template employed for REFIND is illustrated in Figure \ref{fig:REFIND_Prompt} (Appendix \ref{sec:appendix_prompt_details}).
More details for baselines will be discussed in Appendix \ref{sec:appendix_implementation_details}.

\begin{figure*}[ht!]
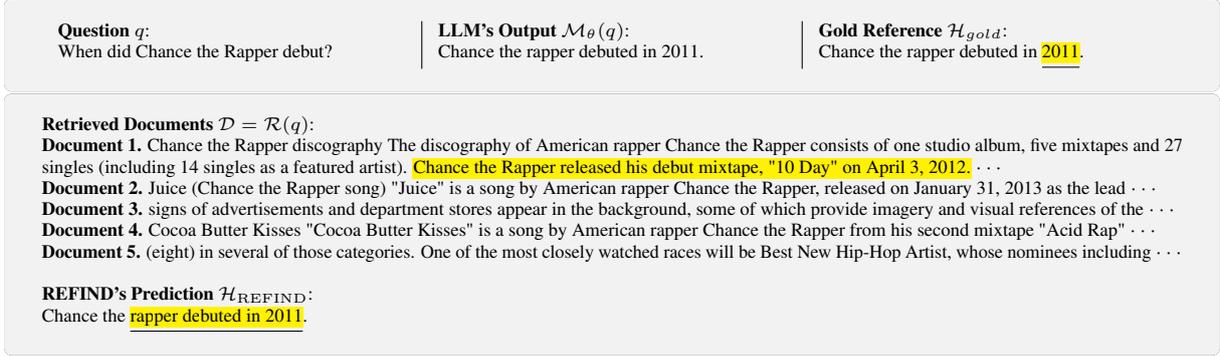

\scriptsize
\begin{tcolorbox}[boxrule=0pt]
    \begin{tabularx}{\textwidth}{X|X|X}
        \textbf{Question} $q$: & \textbf{LLM's Output} $\mathcal{M}_\theta(q)$: & \textbf{Gold Reference} $\mathcal{H}_{gold}$: \\
        When did Chance the Rapper debut? & Chance the rapper debuted in 2011. & Chance the rapper debuted in \underline{\hl{2011}}.
    \end{tabularx}
  % \textbf{Question} $q$: \\
  % When did Chance the Rapper debut?
  % \\\\
  % \textbf{LLM's Output} $\mathcal{M}_\theta(q)$: \\
  % Chance the rapper debuted in 2011.
  % \\\\
  % \textbf{Gold Reference} $\mathcal{H}_{gold}$: \\
  % Chance the rapper debuted in \underline{\hl{2011}}.
\end{tcolorbox}

\vspace{-2.5mm}

\begin{tcolorbox}[boxrule=0pt]

  \textbf{Retrieved Documents} $\mathcal{D}=\mathcal{R}(q)$: \\
  \textbf{Document 1.} Chance the Rapper discography The discography of American rapper Chance the Rapper consists of one studio album, five mixtapes and 27 singles (including 14 singles as a featured artist). \hl{Chance the Rapper released his debut mixtape, "10 Day" on April 3, 2012.} $\cdots$\\
  \textbf{Document 2.} Juice (Chance the Rapper song) "Juice" is a song by American rapper Chance the Rapper, released on January 31, 2013 as the lead $\cdots$\\
  \textbf{Document 3.} signs of advertisements and department stores appear in the background, some of which provide imagery and visual references of the $\cdots$\\
  \textbf{Document 4.} Cocoa Butter Kisses "Cocoa Butter Kisses" is a song by American rapper Chance the Rapper from his second mixtape "Acid Rap" $\cdots$\\
  \textbf{Document 5.} (eight) in several of those categories. One of the most closely watched races will be Best New Hip-Hop Artist, whose nominees including $\cdots$
  \\\\
  \textbf{REFIND's Prediction} $\mathcal{H}_{\mathrm{REFIND}}$: \\
  Chance the \underline{\hl{rapper debuted in 2011}}.
\end{tcolorbox}

\vspace{-3mm}

\caption{Example result of REFIND's hallucination detection. The gold reference $\mathcal{H}_{gold}$ highlights the correct hallucinated span, while REFIND successfully identifies the hallucinated span in the output, demonstrating its alignment with the gold annotations. The complete text of the retrieved documents is available in Appendix \ref{sec:appendix_full_documents}.}

\label{fig:case_study}
\end{figure*}

\section{Result and Analysis}

\subsection{Performance Comparison}

Table \ref{tab:result} presents the evaluation results of our proposed method, alongside the baseline models, XLM-R and FAVA, on the Mu-SHROOM dataset. The results are reported across nine languages (AR, CS, DE, EN, ES, EU, FI, FR, IT) and averaged to provide an overall assessment of performance.

REFIND outperforms the baseline models in terms of average IoU scores.
The improvements are particularly notable in low-resource languages such as Arabic, Finnish, and French, where REFIND achieves IoU scores of 0.3743, 0.5061, and 0.4734, respectively, compared to significantly lower scores from the baselines. This indicates that REFIND effectively leverages retrieval-augmented information to enhance hallucination detection in diverse linguistic settings.

\subsection{Baseline Comparison}

The XLM-R-based token classifier performs poorly on average, with an IoU of 0.0345. Its reliance solely on intrinsic model knowledge without leveraging external context limits its ability to identify hallucinated spans accurately, particularly in low-resource languages.

FAVA exhibits better performance than XLM-R, with an average IoU of 0.2787. This improvement can be attributed to its use of retrieval-augmented information for detecting and editing hallucinated text. However, FAVA's two-step process introduces complexity and potential inaccuracies in aligning the edited text with the original output.

REFIND outperforms both baselines with an average IoU of 0.3633, highlighting its superior ability to integrate retrieved context directly into the token generation process for hallucination detection. 
This streamlined approach ensures accurate and efficient identification of hallucinated spans.

\subsection{Analysis of Multilingual Performance}

REFIND demonstrates robust performance across both high-resource and low-resource languages. This indicates the generalizability of its retrieval-augmented approach to varying linguistic contexts.  Notably, performance varies considerably across languages for all methods; for instance, XLM-R and FAVA struggle significantly with low-resource languages like Arabic, Finnish, and French. In contrast, REFIND's integration of external retrieval with the LLM's internal knowledge helps mitigate performance drops in these settings.

\begin{figure}[t!]
    \centering
    \includegraphics[width=\columnwidth]{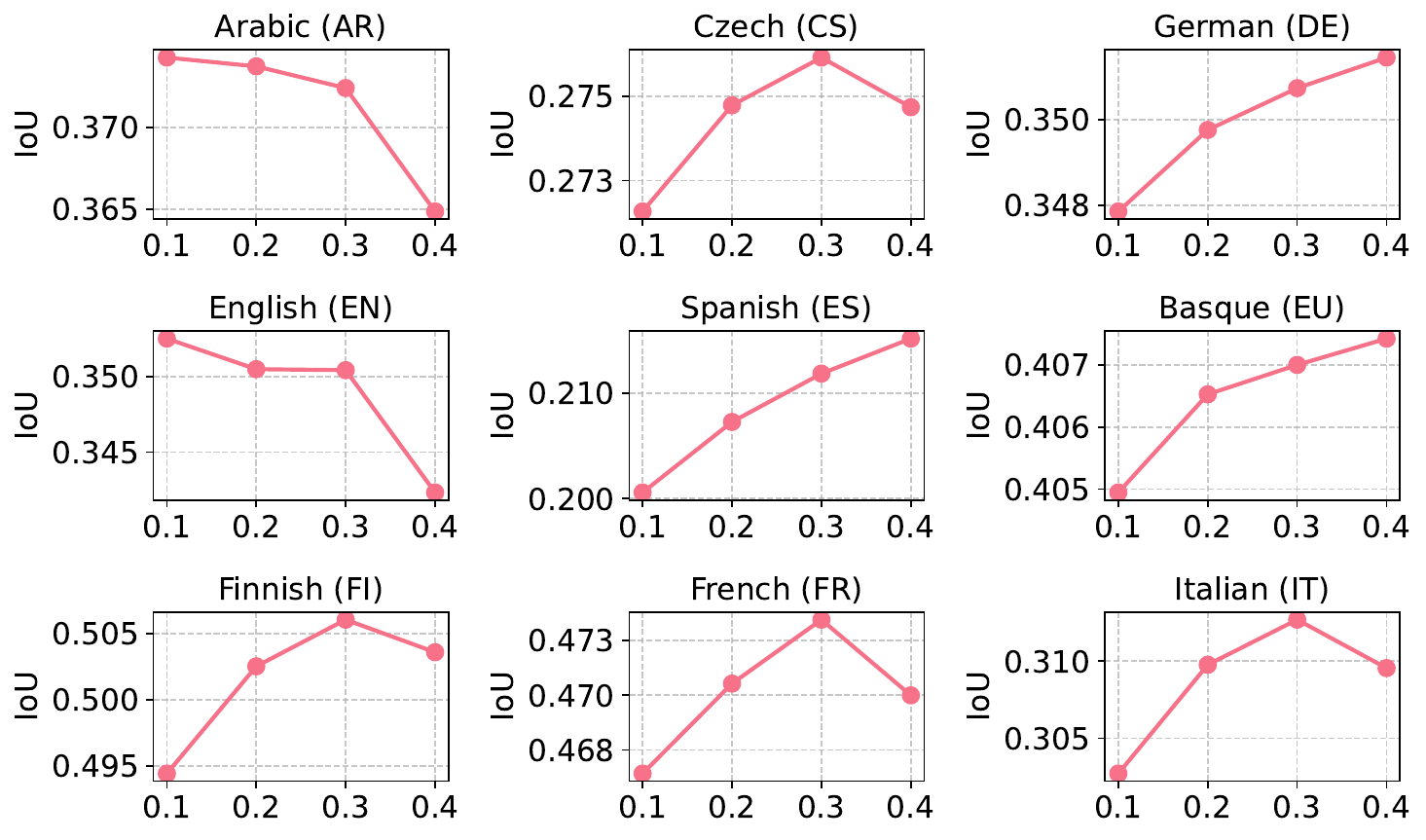}
    \caption {
        Analysis of IoU scores across different threshold values ($\delta \in {0.1, 0.2, 0.3, 0.4}$). Each subplot represents a different language, showing the relationship between threshold values and IoU scores. 
    } 
    \label{fig:threshold_analysis}
\end{figure}

\subsection{Analysis of Threshold Sensitivity}

Figure~\ref{fig:threshold_analysis} illustrates the performance of REFIND across varying threshold values (0.1-0.4) for nine languages.
Most languages exhibit consistent IoU scores, indicating robustness to threshold changes. High-resource languages like English and German maintain stable scores around 0.35, while low-resource languages such as Arabic and Finnish show slightly larger variations, especially at lower thresholds. 
This suggests that the choice of threshold may have a more significant impact on low-resource languages, potentially due to their inherent linguistic challenges and data scarcity. 
Overall, these findings emphasize REFIND's ability to maintain reliable performance across a range of threshold values while highlighting potential areas for optimization in low-resource scenarios.

\subsection{Case Study}
\label{sec:case_study}

Figure~\ref{fig:case_study} illustrates REFIND's ability to detect hallucinations by utilizing retrieved evidence. The question asks about Chance the Rapper's debut year. The LLM's output contains a hallucinated span ("\underline{\hl{2011}}"), which is inconsistent with the retrieved documents. By comparing the generated output with external knowledge, REFIND effectively identifies spans that deviate from factual information. 

\section{Conclusion}
In this study, we introduced REFIND, a novel framework for detecting hallucinated spans in LLM-generated outputs by leveraging retrieved documents to compute the Context Sensitivity Ratio (CSR) at the token level. REFIND was rigorously evaluated on the multilingual SemEval 2025 Task 3: Mu-SHROOM dataset, demonstrating superior performance across nine languages, including low-resource settings, compared to baseline approaches. By directly integrating retrieved context into the token probability calculation, REFIND effectively identifies hallucinated spans with greater precision and efficiency.

Our experimental results highlight the robustness and scalability of REFIND in multilingual environments, offering a promising solution for enhancing the factuality of LLM outputs. Moreover, the streamlined detection process avoids the complexities associated with multi-step frameworks, enabling practical deployment in real-world applications.

For future work, we aim to extend REFIND by exploring adaptive thresholding mechanisms to further optimize the balance between precision and recall in hallucination detection. 
% Additionally, future research is required to investigate how detected hallucinations can guide the decoding process to improve the quality of generated outputs, potentially enabling iterative refinement during generation.

\section*{Limitations}

While REFIND achieves notable improvements in hallucination detection, there are limitations to consider. First, the reliance on retrieved documents means that the quality of the retriever directly impacts performance. Errors in retrieval or limited availability of relevant documents may lead to suboptimal CSR calculations and misclassification of hallucinated spans. Second, the approach involves computational overhead associated with calculating token probabilities with and without retrieved context, which could pose challenges in low-latency applications. Lastly, REFIND focuses on detecting factual hallucinations, and its performance in non-factoid question answering \cite{Bolotova2022NFQATaxanomy, Lee2025Typed-RAG} remains unexplored. Further studies are needed to assess its ability to detect hallucinations in non-factoid QA tasks.

\bibliography{anthology,custom}

\begin{thebibliography}{24}
\providecommand{\natexlab}[1]{#1}

\bibitem[{Ansel et~al.(2024)Ansel, Yang, He, Gimelshein, Jain, Voznesensky, Bao, Bell, Berard, Burovski, Chauhan, Chourdia, Constable, Desmaison, DeVito, Ellison, Feng, Gong, Gschwind, Hirsh, Huang, Kalambarkar, Kirsch, Lazos, Lezcano, Liang, Liang, Lu, Luk, Maher, Pan, Puhrsch, Reso, Saroufim, Siraichi, Suk, Zhang, Suo, Tillet, Zhao, Wang, Zhou, Zou, Wang, Mathews, Wen, Chanan, Wu, and Chintala}]{Jason2024PyTorch2}
Jason Ansel, Edward Yang, Horace He, Natalia Gimelshein, Animesh Jain, Michael Voznesensky, Bin Bao, Peter Bell, David Berard, Evgeni Burovski, Geeta Chauhan, Anjali Chourdia, Will Constable, Alban Desmaison, Zachary DeVito, Elias Ellison, Will Feng, Jiong Gong, Michael Gschwind, Brian Hirsh, Sherlock Huang, Kshiteej Kalambarkar, Laurent Kirsch, Michael Lazos, Mario Lezcano, Yanbo Liang, Jason Liang, Yinghai Lu, C.~K. Luk, Bert Maher, Yunjie Pan, Christian Puhrsch, Matthias Reso, Mark Saroufim, Marcos~Yukio Siraichi, Helen Suk, Shunting Zhang, Michael Suo, Phil Tillet, Xu~Zhao, Eikan Wang, Keren Zhou, Richard Zou, Xiaodong Wang, Ajit Mathews, William Wen, Gregory Chanan, Peng Wu, and Soumith Chintala. 2024.
\newblock \href {https://doi.org/10.1145/3620665.3640366} {Pytorch 2: Faster machine learning through dynamic python bytecode transformation and graph compilation}.
\newblock In \emph{Proceedings of the 29th ACM International Conference on Architectural Support for Programming Languages and Operating Systems, Volume 2}, ASPLOS '24, page 929–947, New York, NY, USA. Association for Computing Machinery.

\bibitem[{Arteaga et~al.(2025)Arteaga, Sch{\"o}n, and Pielawski}]{arteaga2024hallucination}
Gabriel~Y. Arteaga, Thomas~B. Sch{\"o}n, and Nicolas Pielawski. 2025.
\newblock \href {https://openreview.net/forum?id=8T8QkDsuO9} {Hallucination detection in {LLM}s: Fast and memory-efficient finetuned models}.
\newblock In \emph{Northern Lights Deep Learning Conference 2025}.

\bibitem[{Bolotova et~al.(2022)Bolotova, Blinov, Scholer, Croft, and Sanderson}]{Bolotova2022NFQATaxanomy}
Valeriia Bolotova, Vladislav Blinov, Falk Scholer, W.~Bruce Croft, and Mark Sanderson. 2022.
\newblock \href {https://doi.org/10.1145/3477495.3531926} {A non-factoid question-answering taxonomy}.
\newblock In \emph{Proceedings of the 45th International ACM SIGIR Conference on Research and Development in Information Retrieval}, SIGIR '22, page 1196–1207, New York, NY, USA. Association for Computing Machinery.

\bibitem[{Conneau et~al.(2020)Conneau, Khandelwal, Goyal, Chaudhary, Wenzek, Guzm{\'a}n, Grave, Ott, Zettlemoyer, and Stoyanov}]{conneau-etal-2020-unsupervised}
Alexis Conneau, Kartikay Khandelwal, Naman Goyal, Vishrav Chaudhary, Guillaume Wenzek, Francisco Guzm{\'a}n, Edouard Grave, Myle Ott, Luke Zettlemoyer, and Veselin Stoyanov. 2020.
\newblock \href {https://doi.org/10.18653/v1/2020.acl-main.747} {Unsupervised cross-lingual representation learning at scale}.
\newblock In \emph{Proceedings of the 58th Annual Meeting of the Association for Computational Linguistics}, pages 8440--8451, Online. Association for Computational Linguistics.

\bibitem[{Farquhar et~al.(2024)Farquhar, Kossen, Kuhn, and Gal}]{Farquhar2024Detecting}
Sebastian Farquhar, Jannik Kossen, Lorenz Kuhn, and Yarin Gal. 2024.
\newblock \href {https://doi.org/10.1038/s41586-024-07421-0} {Detecting hallucinations in large language models using semantic entropy}.
\newblock \emph{Nature}, 630(8017):625--630.

\bibitem[{Han et~al.(2024)Han, Kossen, Razzak, Schut, Malik, and Gal}]{han2024semantic}
Jiatong Han, Jannik Kossen, Muhammed Razzak, Lisa Schut, Shreshth~A Malik, and Yarin Gal. 2024.
\newblock \href {https://openreview.net/forum?id=Zd0XLr6JKn} {Semantic entropy probes: Robust and cheap hallucination detection in {LLM}s}.
\newblock In \emph{ICML 2024 Workshop on Foundation Models in the Wild}.

\bibitem[{Ji et~al.(2023)Ji, Lee, Frieske, Yu, Su, Xu, Ishii, Bang, Madotto, and Fung}]{Ji2023HallucinationSurvey}
Ziwei Ji, Nayeon Lee, Rita Frieske, Tiezheng Yu, Dan Su, Yan Xu, Etsuko Ishii, Ye~Jin Bang, Andrea Madotto, and Pascale Fung. 2023.
\newblock \href {https://doi.org/10.1145/3571730} {Survey of hallucination in natural language generation}.
\newblock \emph{ACM Computing Surveys}, 55(12).

\bibitem[{Kuhn et~al.(2023)Kuhn, Gal, and Farquhar}]{kuhn2023semantic}
Lorenz Kuhn, Yarin Gal, and Sebastian Farquhar. 2023.
\newblock \href {https://openreview.net/forum?id=VD-AYtP0dve} {Semantic uncertainty: Linguistic invariances for uncertainty estimation in natural language generation}.
\newblock In \emph{The Eleventh International Conference on Learning Representations}.

\bibitem[{Kwon et~al.(2023)Kwon, Li, Zhuang, Sheng, Zheng, Yu, Gonzalez, Zhang, and Stoica}]{Kwon2023vLLM}
Woosuk Kwon, Zhuohan Li, Siyuan Zhuang, Ying Sheng, Lianmin Zheng, Cody~Hao Yu, Joseph Gonzalez, Hao Zhang, and Ion Stoica. 2023.
\newblock \href {https://doi.org/10.1145/3600006.3613165} {Efficient memory management for large language model serving with pagedattention}.
\newblock In \emph{Proceedings of the 29th Symposium on Operating Systems Principles}, SOSP '23, page 611–626, New York, NY, USA. Association for Computing Machinery.

\bibitem[{Lee et~al.(2025)Lee, Park, Lee, Nam, and Maeng}]{Lee2025Typed-RAG}
DongGeon Lee, Ahjeong Park, Hyeri Lee, Hyeonseo Nam, and Yunho Maeng. 2025.
\newblock \href {https://arxiv.org/abs/2503.15879} {{Typed-RAG}: {T}ype-aware multi-aspect decomposition for non-factoid question answering}.
\newblock \emph{arXiv preprint arXiv:2503.15879}.

\bibitem[{Lee et~al.(2022)Lee, Ping, Xu, Patwary, Fung, Shoeybi, and Catanzaro}]{Lee2022Factuality}
Nayeon Lee, Wei Ping, Peng Xu, Mostofa Patwary, Pascale~N Fung, Mohammad Shoeybi, and Bryan Catanzaro. 2022.
\newblock \href {https://proceedings.neurips.cc/paper_files/paper/2022/file/df438caa36714f69277daa92d608dd63-Paper-Conference.pdf} {Factuality enhanced language models for open-ended text generation}.
\newblock In \emph{Advances in Neural Information Processing Systems}, volume~35, pages 34586--34599. Curran Associates, Inc.

\bibitem[{Li et~al.(2024)Li, Chen, Ren, Cheng, Zhao, Nie, and Wen}]{li-etal-2024-dawn}
Junyi Li, Jie Chen, Ruiyang Ren, Xiaoxue Cheng, Xin Zhao, Jian-Yun Nie, and Ji-Rong Wen. 2024.
\newblock \href {https://doi.org/10.18653/v1/2024.acl-long.586} {The dawn after the dark: An empirical study on factuality hallucination in large language models}.
\newblock In \emph{Proceedings of the 62nd Annual Meeting of the Association for Computational Linguistics (Volume 1: Long Papers)}, pages 10879--10899, Bangkok, Thailand. Association for Computational Linguistics.

\bibitem[{Liu et~al.(2022)Liu, Zhang, Brockett, Mao, Sui, Chen, and Dolan}]{liu-etal-2022-token}
Tianyu Liu, Yizhe Zhang, Chris Brockett, Yi~Mao, Zhifang Sui, Weizhu Chen, and Bill Dolan. 2022.
\newblock \href {https://doi.org/10.18653/v1/2022.acl-long.464} {A token-level reference-free hallucination detection benchmark for free-form text generation}.
\newblock In \emph{Proceedings of the 60th Annual Meeting of the Association for Computational Linguistics (Volume 1: Long Papers)}, pages 6723--6737, Dublin, Ireland. Association for Computational Linguistics.

\bibitem[{Liu et~al.(2019)Liu, Ott, Goyal, Du, Joshi, Chen, Levy, Lewis, Zettlemoyer, and Stoyanov}]{Liu2019RoBERTa}
Yinhan Liu, Myle Ott, Naman Goyal, Jingfei Du, Mandar Joshi, Danqi Chen, Omer Levy, Mike Lewis, Luke Zettlemoyer, and Veselin Stoyanov. 2019.
\newblock \href {https://arxiv.org/abs/1907.11692} {Roberta: {A} robustly optimized {BERT} pretraining approach}.
\newblock \emph{arXiv preprint arXiv:1907.11692}.

\bibitem[{Manakul et~al.(2023)Manakul, Liusie, and Gales}]{manakul-etal-2023-selfcheckgpt}
Potsawee Manakul, Adian Liusie, and Mark Gales. 2023.
\newblock \href {https://doi.org/10.18653/v1/2023.emnlp-main.557} {{S}elf{C}heck{GPT}: Zero-resource black-box hallucination detection for generative large language models}.
\newblock In \emph{Proceedings of the 2023 Conference on Empirical Methods in Natural Language Processing}, pages 9004--9017, Singapore. Association for Computational Linguistics.

\bibitem[{Mishra et~al.(2024)Mishra, Asai, Balachandran, Wang, Neubig, Tsvetkov, and Hajishirzi}]{mishra2024finegrained-FAVA}
Abhika Mishra, Akari Asai, Vidhisha Balachandran, Yizhong Wang, Graham Neubig, Yulia Tsvetkov, and Hannaneh Hajishirzi. 2024.
\newblock \href {https://openreview.net/forum?id=dJMTn3QOWO} {Fine-grained hallucination detection and editing for language models}.
\newblock In \emph{The First Conference on Language Modeling}.

\bibitem[{Robertson and Zaragoza(2009)}]{Robertson2009BM25}
Stephen Robertson and Hugo Zaragoza. 2009.
\newblock \href {https://doi.org/10.1561/1500000019} {The probabilistic relevance framework: Bm25 and beyond}.
\newblock \emph{Foundations and Trends in Information Retrieval}, 3(4):333–389.

\bibitem[{Sun et~al.(2024)Sun, Cai, Wang, Hou, Wei, Wang, Zhang, and Yin}]{sun-etal-2024-towards-verifiable}
Hao Sun, Hengyi Cai, Bo~Wang, Yingyan Hou, Xiaochi Wei, Shuaiqiang Wang, Yan Zhang, and Dawei Yin. 2024.
\newblock \href {https://doi.org/10.18653/v1/2024.emnlp-main.469} {Towards verifiable text generation with evolving memory and self-reflection}.
\newblock In \emph{Proceedings of the 2024 Conference on Empirical Methods in Natural Language Processing}, pages 8211--8227, Miami, Florida, USA. Association for Computational Linguistics.

\bibitem[{Touvron et~al.(2023)Touvron, Martin, Stone, Albert, Almahairi, Babaei, Bashlykov, Batra, Bhargava, Bhosale, Bikel, Blecher, Canton{-}Ferrer, Chen, Cucurull, Esiobu, Fernandes, Fu, Fu, Fuller, Gao, Goswami, Goyal, Hartshorn, Hosseini, Hou, Inan, Kardas, Kerkez, Khabsa, Kloumann, Korenev, Koura, Lachaux, Lavril, Lee, Liskovich, Lu, Mao, Martinet, Mihaylov, Mishra, Molybog, Nie, Poulton, Reizenstein, Rungta, Saladi, Schelten, Silva, Smith, Subramanian, Tan, Tang, Taylor, Williams, Kuan, Xu, Yan, Zarov, Zhang, Fan, Kambadur, Narang, Rodriguez, Stojnic, Edunov, and Scialom}]{Touvron2023Llama2}
Hugo Touvron, Louis Martin, Kevin Stone, Peter Albert, Amjad Almahairi, Yasmine Babaei, Nikolay Bashlykov, Soumya Batra, Prajjwal Bhargava, Shruti Bhosale, Dan Bikel, Lukas Blecher, Cristian Canton{-}Ferrer, Moya Chen, Guillem Cucurull, David Esiobu, Jude Fernandes, Jeremy Fu, Wenyin Fu, Brian Fuller, Cynthia Gao, Vedanuj Goswami, Naman Goyal, Anthony Hartshorn, Saghar Hosseini, Rui Hou, Hakan Inan, Marcin Kardas, Viktor Kerkez, Madian Khabsa, Isabel Kloumann, Artem Korenev, Punit~Singh Koura, Marie{-}Anne Lachaux, Thibaut Lavril, Jenya Lee, Diana Liskovich, Yinghai Lu, Yuning Mao, Xavier Martinet, Todor Mihaylov, Pushkar Mishra, Igor Molybog, Yixin Nie, Andrew Poulton, Jeremy Reizenstein, Rashi Rungta, Kalyan Saladi, Alan Schelten, Ruan Silva, Eric~Michael Smith, Ranjan Subramanian, Xiaoqing~Ellen Tan, Binh Tang, Ross Taylor, Adina Williams, Jian~Xiang Kuan, Puxin Xu, Zheng Yan, Iliyan Zarov, Yuchen Zhang, Angela Fan, Melanie Kambadur, Sharan Narang, Aur{\'{e}}lien Rodriguez, Robert Stojnic, Sergey Edunov,
  and Thomas Scialom. 2023.
\newblock \href {https://doi.org/10.48550/ARXIV.2307.09288} {Llama 2: Open foundation and fine-tuned chat models}.
\newblock \emph{arXiv preprint arXiv:2307.09288}.

\bibitem[{V\'azquez et~al.(2025)V\'azquez, Mickus, Zosa, Vahtola, Tiedemann, Sinha, Segonne, S\'anchez-Vega, Raganato, Libovický, Karlgren, Ji, Helcl, Guillou, de~Gibert, Bengoetxea, Attieh, and Apidianaki}]{vazquez-etal-2025-mu-shroom}
Ra\'ul V\'azquez, Timothee Mickus, Elaine Zosa, Teemu Vahtola, J\"org Tiedemann, Aman Sinha, Vincent Segonne, Fernando S\'anchez-Vega, Alessandro Raganato, Jindřich Libovický, Jussi Karlgren, Shaoxiong Ji, Jindřich Helcl, Liane Guillou, Ona de~Gibert, Jaione Bengoetxea, Joseph Attieh, and Marianna Apidianaki. 2025.
\newblock \href {https://helsinki-nlp.github.io/shroom/} {Sem{E}val-2025 {T}ask 3: {Mu-SHROOM}, the multilingual shared-task on hallucinations and related observable overgeneration mistakes}.

\bibitem[{Wang et~al.(2024)Wang, Yang, Huang, Yang, Majumder, and Wei}]{Wang2024multilingual-e5}
Liang Wang, Nan Yang, Xiaolong Huang, Linjun Yang, Rangan Majumder, and Furu Wei. 2024.
\newblock \href {https://doi.org/10.48550/ARXIV.2402.05672} {Multilingual {E5} text embeddings: {A} technical report}.
\newblock \emph{arXiv preprint arXiv:2402.05672}.

\bibitem[{Wolf et~al.(2020)Wolf, Debut, Sanh, Chaumond, Delangue, Moi, Cistac, Rault, Louf, Funtowicz, Davison, Shleifer, von Platen, Ma, Jernite, Plu, Xu, Le~Scao, Gugger, Drame, Lhoest, and Rush}]{wolf-etal-2020-transformers}
Thomas Wolf, Lysandre Debut, Victor Sanh, Julien Chaumond, Clement Delangue, Anthony Moi, Pierric Cistac, Tim Rault, Remi Louf, Morgan Funtowicz, Joe Davison, Sam Shleifer, Patrick von Platen, Clara Ma, Yacine Jernite, Julien Plu, Canwen Xu, Teven Le~Scao, Sylvain Gugger, Mariama Drame, Quentin Lhoest, and Alexander Rush. 2020.
\newblock \href {https://doi.org/10.18653/v1/2020.emnlp-demos.6} {Transformers: State-of-the-art natural language processing}.
\newblock In \emph{Proceedings of the 2020 Conference on Empirical Methods in Natural Language Processing: System Demonstrations}, pages 38--45, Online. Association for Computational Linguistics.

\bibitem[{Yuksekgonul et~al.(2024)Yuksekgonul, Chandrasekaran, Jones, Gunasekar, Naik, Palangi, Kamar, and Nushi}]{yuksekgonul2024attention}
Mert Yuksekgonul, Varun Chandrasekaran, Erik Jones, Suriya Gunasekar, Ranjita Naik, Hamid Palangi, Ece Kamar, and Besmira Nushi. 2024.
\newblock \href {https://openreview.net/forum?id=gfFVATffPd} {Attention satisfies: A constraint-satisfaction lens on factual errors of language models}.
\newblock In \emph{The Twelfth International Conference on Learning Representations}.

\bibitem[{Zhang et~al.(2023)Zhang, Li, Cui, Cai, Liu, Fu, Huang, Zhao, Zhang, Chen, Wang, Luu, Bi, Shi, and Shi}]{Zhang2023Sirens_Song_Hallucination}
Yue Zhang, Yafu Li, Leyang Cui, Deng Cai, Lemao Liu, Tingchen Fu, Xinting Huang, Enbo Zhao, Yu~Zhang, Yulong Chen, Longyue Wang, Anh~Tuan Luu, Wei Bi, Freda Shi, and Shuming Shi. 2023.
\newblock \href {https://doi.org/10.48550/ARXIV.2309.01219} {Siren's song in the {AI} ocean: {A} survey on hallucination in large language models}.
\newblock \emph{arXiv preprint arXiv:2309.01219}.

\end{thebibliography}

\appendix
\clearpage
\onecolumn

\section{Implementation Details}
\label{sec:appendix_implementation_details}
All experiments are conducted using NVIDIA A100 80GB GPUs. 

For training the XLM-R-based \cite{conneau-etal-2020-unsupervised} system, we leverage the Trainer from the Hugging Face Transformers library \cite{wolf-etal-2020-transformers}. 
We train the model using token-aligned hallucination annotations from our dataset, with the model parameters optimized using cross-entropy loss and AdamW optimizer with a learning rate of 2e-5 for 5 epochs.

Inference for FAVA \cite{mishra2024finegrained-FAVA} is conducted using vLLM \cite{Kwon2023vLLM}, adhering to the original settings with \textit{temperature}=0, \textit{top\_p}=1.0, and \textit{max\_tokens}=1024. The prompt template used for FAVA inference is detailed in Figure \ref{fig:FAVA_Prompt} (Appendix \ref{sec:appendix_prompt_details}).

\subsection{Prompt Details}
\label{sec:appendix_prompt_details}

\begin{figure}[htb!]
    \centering
    \begin{tcolorbox}[colback=gray!10, colframe=black, title=Prompt template for REFIND]
        You are an assistant for answering questions.\\
        Refer to the references below and answer the following question.
        \\ \\
        \#\#\# References\\
        \{\texttt{reference\_passages}\} \\ \\
        \#\#\# Question\\
        \{\texttt{question}\} \\ \\
        \#\#\# Answer
    \end{tcolorbox}
    \vspace{-3mm}
    
    \caption{Prompt template of REFIND used to compute per-token probabilities under the conditions provided in the input context.}
    \label{fig:REFIND_Prompt}
\end{figure}

\begin{figure}[htb!]
    \centering
    \begin{tcolorbox}[colback=gray!10, colframe=black, title=Prompt template for FAVA]
        Read the following references:\\
        \{\texttt{reference\_passages}\} \\
        Please identify all the errors in the following text using the information in the references provided and suggest edits if necessary:\\
        \lbrack Text\rbrack\ \{\texttt{output}\}\\
        \lbrack Edited\rbrack\ 
    \end{tcolorbox}
    \vspace{-3mm}
    
    \caption{Prompt template for using FAVA \cite{mishra2024finegrained-FAVA}.}
    \label{fig:FAVA_Prompt}
\end{figure}

\onecolumn
\section{Full Text of Retrieved Documents \texorpdfstring{$\mathcal{D}$}{D} for Case Study\texorpdfstring{ (\cref{sec:case_study})}{}}
\label{sec:appendix_full_documents}

\begin{figure*}[ht!]

\large
\begin{tcolorbox}[boxrule=0pt]
  \textbf{Document 1.} Chance the Rapper discography he discography of American rapper Chance the Rapper consists of one studio album, five mixtapes and 27 singles (including 14 singles as a featured artist). Chance the Rapper released his debut mixtape, "10 Day" on April 3, 2012. The mixtape was followed up with the release of "Acid Rap" the following year, which saw universal acclaim from music critics. Chance the Rapper then released his third mixtape, "Coloring Book" on May 13, 2016. The mixtape peaked at number eight on the "Billboard" 200 chart to continued acclaim and was supported by the singles "Angels"\\\\
  \textbf{Document 2.} Juice (Chance the Rapper song) "Juice" is a song by American rapper Chance the Rapper, released on January 31, 2013 as the lead single from his second mixtape "Acid Rap" (2013). It was written by Chance and Nate Fox, who also produced the song. "Juice" is a midtempo song, built around a loop of Donny Hathaway's live performance of "Jealous Guy" by John Lennon. Chance the Rapper sings and raps in a comedic manner; his verses in the song have been described as having a "freewheeling, bluesy sway" that "gives way to raucous call-and-response choruses". He references the 1992 film "Juice" (of\\\\
  \textbf{Document 3.} signs of advertisements and department stores appear in the background, some of which provide imagery and visual references of the lyrics. For example, when Chance lyrically alludes to the film "Juice", a portrait of rapper Tupac Shakur (who starred in the film) flashes across a billboard. When "Acid Rap" was first re-released on streaming services on June 28, 2019, "Juice" was replaced with a 30-second spoken message, in which Chance the Rapper explains the song is excluded from the mixtape because of an uncleared sample. Chance then adds that all streaming proceeds for the alternate\\\\
  \textbf{Document 4.} Cocoa Butter Kisses "Cocoa Butter Kisses" is a song by American rapper Chance the Rapper from his second mixtape "Acid Rap" (2013). The song features American rappers Vic Mensa and Twista, and was produced by Cam O'bi and Peter Cottontale. It is one of Chance the Rapper's most popular songs to date. At the time when the song was written, Vic Mensa was staying at an apartment in Humboldt Park, Chicago with his manager Cody Kazarian. Chance the Rapper visited one day and showed Mensa a verse and hook he had written earlier. Soon, Mensa began composing his part for the song. In an interview\\\\
  \textbf{Document 5.} (eight) in several of those categories. One of the most closely watched races will be Best New Hip-Hop Artist, whose nominees including Anderson .Paak, Bryson Tiller (who won that award and Best Male R\&B/Pop Artist at June’s BET Awards), Chance the Rapper, Desiigner and Tory Lanez.  Drake – "Hotline Bling" Fat Joe \& Remy Ma featuring French Montana \& Infared – "All the Way Up" Kendrick Lamar Kendrick Lamar Director X DJ Khaled Metro Boomin DJ Khaled "All the Way Up" – Produced by Cool \& Dre and Edsclusive  Drake – "Views" Chance the Rapper DJ Khaled Kanye West Chance the Rapper

\end{tcolorbox}

\vspace{-3mm}

\caption{Complete text of documents retrieved for the input question "\textit{When did Chance the Rapper debut?}" as referenced in the case study in Section \ref{sec:case_study}.}

\label{fig:full_documents}
\end{figure*}

\end{document}